\documentclass[final]{cvpr}
\pdfoutput=1
\usepackage{multicol}
\usepackage{subfig}
\usepackage{times}
\usepackage{epsfig}
\usepackage{graphicx}
\usepackage{amsmath}
\usepackage{amssymb}
\usepackage{graphicx}
\usepackage{subfig}
\usepackage{siunitx}
\usepackage{adjustbox}
\usepackage{multirow}
\usepackage[ruled,vlined]{algorithm2e}
\usepackage[noend]{algpseudocode}
\usepackage[T1]{fontenc} 
\usepackage[pagebackref=true,breaklinks=true,colorlinks,bookmarks=false]{hyperref}

\def\cvprPaperID{5931} 
\def\confYear{CVPR 2021}

\begin{document}

\title{Augmentation Strategies for Learning with Noisy Labels}

\author{
Kento Nishi \thanks{Equal contribution} \! \footnote ,  \quad Yi Ding\footnotemark[1] \! \footnote , \quad Alex Rich \footnotemark[3]  \quad Tobias H\"{o}llerer \footnotemark[3]\\ \\

\footnotemark[2] \! Lynbrook High School, San Jose CA, USA\\
\footnotemark[3] \! University of California Santa Barbara, Santa Barbara CA, USA\\
{\tt\small kento24gs@outlook.com, yding@cs.ucsb.edu, anrich@cs.ucsb.edu, holl@cs.ucsb.edu}
}

\maketitle

\begin{abstract}
Imperfect labels are ubiquitous in real-world datasets. Several recent successful methods for training deep neural networks (DNNs) robust to label noise have used two primary techniques: filtering samples based on loss during a warm-up phase to curate an initial set of cleanly labeled samples, and using the output of a network as a pseudo-label for subsequent loss calculations. 
In this paper, we evaluate different augmentation strategies for algorithms tackling the "learning with noisy labels" problem. We propose and examine multiple augmentation strategies and evaluate them using synthetic datasets based on CIFAR-10 and CIFAR-100, as well as on the real-world dataset Clothing1M. 
Due to several commonalities in these algorithms, we find that using one set of augmentations for loss modeling tasks and another set for learning is the most effective, improving results on the state-of-the-art and other previous methods. Furthermore, we find that applying augmentation during the warm-up period can negatively impact the loss convergence behavior of correctly versus incorrectly labeled samples. We introduce this augmentation strategy to the state-of-the-art technique and demonstrate that we can improve performance across all evaluated noise levels. In particular, we improve accuracy on the CIFAR-10 benchmark at 90\% symmetric noise by more than 15\% in absolute accuracy, and we also improve performance on the Clothing1M dataset.

\end{abstract}

\let\thefootnote\relax\footnotetext{ \\ Source code is available at \url{https://github.com/KentoNishi/Augmentation-for-LNL}.}

\section{Introduction}

Data augmentation is a common method used to expand datasets and has been applied successfully in many computer vision problems such as image classification \cite{xie2020self} and object detection \cite{sohn2020simple}, among many others. In particular, there has been much success using learned augmentations such as AutoAugment \cite{cubuk2019autoaugment} and RandAugment \cite{cubuk2020randaugment} which do not require an expert who knows the dataset to curate augmentation policies. It has been shown that incorporating augmentation policies during training can improve generalization and robustness \cite{hendrycks2019augmix, devries2017improved}. However, few works have explored their efficacy for the domain of learning with noisy labels (LNL) \cite{natarajan2013learning}.

Many techniques which tackle the LNL problem make use of the network memorization effect, where correctly labeled data fit before incorrectly labeled data as discovered by Arpit et al. \cite{arpit2017closer}. This phenomenon was successfully explored in Deep Neural Networks (DNNs) through modeling the loss function and the training process, leading to the development of approaches such as loss correction \cite{tanaka2018joint} and sample selection \cite{han2018coteaching}. Recently, the incorporation of MixUp augmentation \cite{zhang2017mixup} has dramatically improved the ability for algorithms to tolerate higher noise levels \cite{arazo2019unsupervised, li2020dividemix}.

While many existing works use the common random flip and crop image augmentation which we refer to as \textit{weak augmentation}, to the best of our knowledge, no work at the time of writing has explored using more aggressive augmentation from learned policies such as AutoAugment during training for LNL algorithms. These stronger augmentation policies include transformations such as rotate, invert, sheer, etc. We propose to incorporate these stronger augmentation policies into existing architectures in a strategic way to improve performance. Our intuition is that for any augmentation technique to succeed, they must \textit{(1)} improve the generalization of the dataset and \textit{(2)} not negatively impact the loss modeling and loss convergence behavior that LNL techniques rely on. 

With this in mind, we propose an augmentation strategy we call Augmented Descent (\textsc{AugDesc}) to benefit from data augmentation without negatively impacting the network memorization effect. Our idea for \textsc{AugDesc} is to use two different augmentations: a weak augmentation for any loss modeling and pseudo-labeling task, and a strong augmentation for the back-propagation step to improve generalization.

In this paper, we propose and examine how we can incorporate stronger augmentation into existing LNL algorithms to yield improved results. We provide some answers to this problem through the following contributions:

\begin{itemize}
    \item We propose an augmentation strategy, Augmented Descent, which demonstrates state-of-the-art performance on synthetic and real-world datasets under noisy label scenarios. We show empirically that this can increase performance across all evaluated noise levels (Section \ref{sec:synthetic}). In particular, we improve accuracy on the CIFAR-10 benchmark at 90\% symmetric noise by more than 15\% in absolute accuracy, and we also improve performance on the real-world dataset Clothing1M (Section \ref{sec:clo1m}).
    \item We show that there is a large effect on performance depending on how augmentation is incorporated into the training process (Section \ref{sec:aug-strat}). We empirically determine that it is best to use weaker augmentation during earlier epochs followed by stronger augmentations to not adversely affect the memorization effect. We analyze the behavior of loss distribution to yield insight to guide effective incorporation of augmentation in future work (Section \ref{sec:warmup}).
    \item We evaluate the effectiveness of our augmentation methodology by performing generalization studies on existing techniques (Section \ref{sec:generalization}). Without tuning any hyperparameters, we were able to improve existing techniques with only the addition of our proposed augmentation strategy by up to 5\% in absolute accuracy.
\end{itemize}

\section{Related Work}

\textbf{Learning with Noisy Labels}
The most recent advances in training with noisy labels use varying strategies of \textit{(1)} selecting or heavily weighting a subset of clean labels during training \cite{malach2017decoupling, jiang2018mentornet, han2018coteaching, chen2019understanding}, or \textit{(2)} using the output predictions of the DNN or an additional network to correct the loss \cite{Reed2014bootstrap, patrini2017forward, goldberger2017smodel, tanaka2018joint, ma2018dimensionalitydriven}.

Many methods use varying strategies of training two networks, using the output of one or both networks to guide selection of inputs with clean labels.
Decoupling \cite{malach2017decoupling} maintains two networks during training, updating their parameters using only inputs which the two networks disagree on.
MentorNet \cite{jiang2018mentornet} pre-trains an extra network and uses the pre-trained network to apply weights to cleanly labeled inputs more heavily during training of a student network.
Co-teaching \cite{han2018coteaching} maintains two networks, and feeds the low-loss inputs of each network to its peer for parameter updating.
The low-loss inputs are expected to be clean, following the finding that DNNs fit to the underlying clean distribution before overfitting to noisy labels \cite{arpit2017closer}.
INCV \cite{chen2019understanding} trains two networks on mutually exclusive partitions of the training dataset, then uses cross-validation to select clean inputs.
INCV uses the Co-teaching architecture for its networks.
The main drawback of these strategies is they only utilize a subset of the information available for training.

The second category of techniques attempts to use the model's output prediction to correct the loss at training time.
One such common method is to estimate the noise transition matrix and use it to correct the loss, as in forward and backward correction \cite{patrini2017forward} and S-Model \cite{goldberger2017smodel}.
Another common method is to linearly combine the output of the network and the noisy label for calculating loss.
Bootstrap \cite{Reed2014bootstrap} replaces labels with a combination of the label and the prediction from the DNN.
Joint Optimization \cite{tanaka2018joint} uses a similar approach to the work in \cite{Reed2014bootstrap}, but adds a term to the loss to optimize the correction of noisy labels.
D2L \cite{ma2018dimensionalitydriven} monitors the dimensionality of subspaces during training, using it to guide weighting of a linear combination of output prediction and noisy label during loss calculation.

\textbf{Optimized Augmentation}
Augmentation of training data is a widely used method for improving generalization of machine learning models.
Recent works such as AutoAugment \cite{cubuk2019autoaugment} and RandAugment \cite{cubuk2020randaugment} have focused on studying which augmentation policies are optimal.
AutoAugment uses reinforcement learning to determine the selection and ordering of a set of augmentation functions in order to optimize validation loss.
To remove the search phase of AutoAugment and therefore reduce training complexity, RandAugment drastically reduces the search space for optimal augmentations and uses grid search to determine the optimal set.
Both techniques are widely used in semi-supervised settings. 

In semi-supervised learning settings, augmentation has been successfully applied to consistency regularization \cite{sajjadi2016consistency, xie2020uda, berthelot2020remix, sohn2020fixmatch}.
In consistency regularization, a loss is applied to minimize the difference in network prediction between two versions of the same input during training.
\cite{sajjadi2016consistency} uses a mixture of augmentation, random dropout, and random max-pooling to produce these two versions.
More recently, unsupervised data augmentation \cite{xie2020uda} and ReMixMatch \cite{berthelot2020remix} minimize the network predictions between a strongly augmented and weakly augmented version of the input. All of these findings motivate us to incorporate strong augmentation within the realm of LNL to improve performance.

The semi-supervised learning problem itself is similar to the LNL problem with the subtle difference that some labels are unknown rather than corrupt. As techniques in semi-supervised learning have been able to make predictions on a larger dataset from a smaller clean dataset, it would be logical that LNL techniques would benefit from the generalization effects of augmentation. In fact, the recent semi-supervised techniques MixUp \cite{zhang2017mixup}, and Luo et al. \cite{luo2018smooth} all exhibit strong robustness to label noise.


Most recently, FixMatch \cite{sohn2020fixmatch} successfully combines strong vs. weak augmentation in consistency regularization with pseudo-labeling to achieve state-of-the-art results in semi-supervised classification tasks. 
While we similarly employ two separate pools of augmentation functions for use in downstream tasks, there are key important differences.
Most notably, our key idea is separating augmentations used during loss analysis from augmentations used during back-propagation, rather than focusing on pseudo-labeling and consistency regularization.
Additionally, we apply this idea to LNL, a separate domain with different considerations. We experimentally show improvements for a wide variety of LNL algorithms and demonstrate improvements on both synthetic and real-world datasets.

\begin{figure*}[t]
    \begin{minipage}[c]{\textwidth}
        \begin{minipage}[t]{0.23\textwidth}
            \centering
            \subfloat[Raw]{\includegraphics[height = 1in]{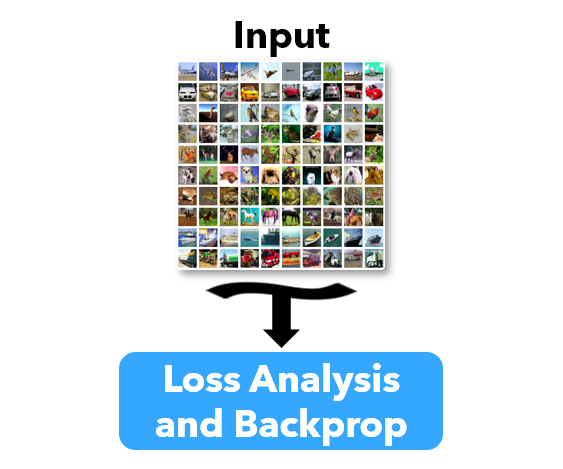}}
        \end{minipage}
        \hfill
        \begin{minipage}[t]{0.23\textwidth}
            \centering
            \subfloat[Dataset Expansion]{\includegraphics[height = 1in]{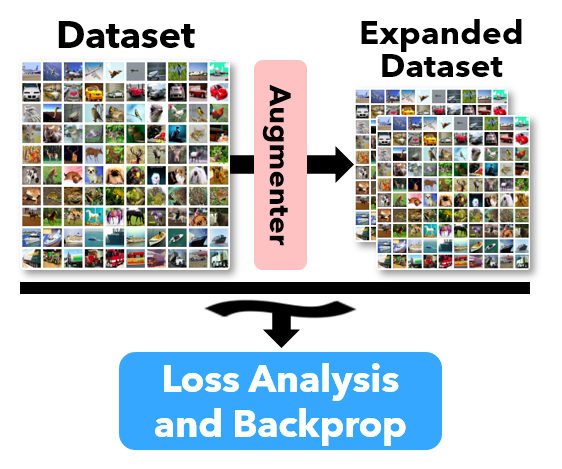}}
        \end{minipage}
        \hfill
        \begin{minipage}[t]{0.23\textwidth}
            \centering
            \subfloat[Runtime]{\includegraphics[height = 1in]{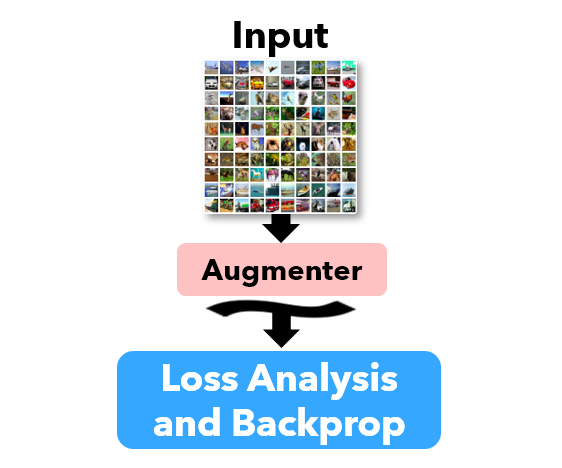}}
        \end{minipage}
        \hfill
        \begin{minipage}[t]{0.23\textwidth}
            \centering
            \subfloat[Augmented Descent]{\includegraphics[height = 1in]{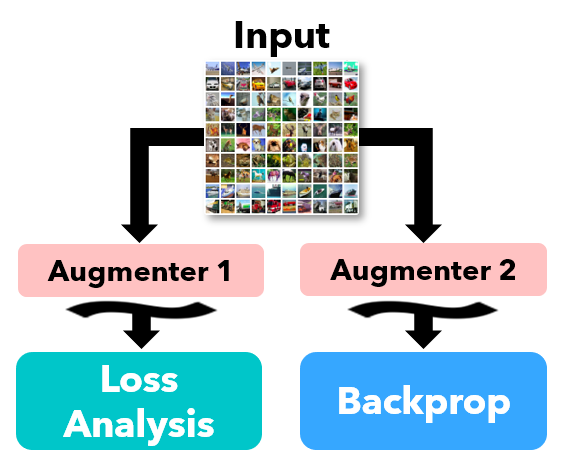}}
        \end{minipage}
        \hfill
    \end{minipage}
\caption{Visualization of training methods when incorporating different augmentation strategies. Raw takes the input directly and feeds it into the model for loss analysis and back-propagation. Dataset expansion first creates an expanded dataset which is then sampled by batches and fed into the network. Runtime Augmentation applies a random augmentation policy during runtime for each sampled batch. Augmented Descent produces two sets of random augmentations at the batch level: one is used for all loss analysis tasks, and the other is used for gradient descent.}
\label{fig:strat}
\end{figure*}

\section{Method}

We first describe how various algorithms operate within the context of the network memorization effect \cite{arpit2017closer}. We then propose the Augmented Descent strategy for filtering and generating pseudo-labels for high confidence samples based on one set of augmentations, then performing gradient descent on a different set of augmentations. Lastly, we provide an example for how to retrofit existing techniques. 

\subsection{Loss Modeling Under Noisy Label Scenarios}

For some training data $D = {(x_i, y_i)}^N_{i = 1}$, a classifier can be trained to make predictions using the cross entropy loss:

\[
l(\theta) = -\sum_{x, y \in D} y^T \log (h_\theta(x)),
\]
where $h_\theta$ is the function approximated by a neural network. Fundamentally, many algorithms are exploiting the behavior outlined in Arpit et al. \cite{arpit2017closer} which finds that correctly labeled data tends to converge before incorrectly label data when training neural networks.

Many existing algorithms are then employing some degree of "pseudo-labeling", where the network is using its own guesses to approximate the labels for the remainder of the dataset. This is done by encouraging the learning of high confidence (or lower initial loss) samples via filtering or modifications to the loss function. 

For example, in the sample selection technique Co-teaching \cite{han2018coteaching}, this is accomplished by feeding low-loss samples to a sister network, training the networks on data which it believes is correct. Abstractly, this would create two datasets from the input for each training epoch of what is believed to be correctly labeled $C = arg\,min_{D:|D| \geq R(T)|D|} l(f, D)$, where $R(T)$ is a threshold for the number of samples to place into the clean set determined empirically by the loss behavior, and incorrectly labeled $I = D \setminus C$. Using these sets, we obtain the loss:

\begin{align*}
l(\theta) = -\sum_{x, y \in C} y^T \log (h_\theta(x)) - 0 * \sum_{x, y \in I} y^T \log (h_\theta(x)).
\end{align*}
Here, the learning process is ignoring samples which are believed to be incorrectly labeled as the training progresses. This is represented by the $0$ term multiplied into what the model believes to be incorrect samples.

By contrast, Arazo et al. \cite{arazo2019unsupervised} accomplishes noise tolerance by incorporating the network's own prediction into its loss as a weighted sum based on the confidence determined by a mixture model fit to the previous epoch's losses, enabling a softer incorporation of the labels:

\begin{align*}
    l(\theta) = - & \sum_{x, y \in D, w \in W} (1 - w) y^T \log (h_\theta(x))  \\
    &- \sum_{x \in D, w \in W} w z^T \log (h_\theta(x)), 
\end{align*}
where $W$ is a set of weights learned using a beta mixture model and $z$ is the model's prediction for input $x$. More recently, DivideMix \cite{li2020dividemix} combines these ideas and assigns weights to inputs to incorporate network guesses, separates the input into two sets, and trains with the resulting data in a semi-supervised manner using MixMatch \cite{berthelot2019mixmatch}.

With this understanding, we propose Augmented Descent (\textsc{AugDesc}) for LNL techniques that employ loss modeling to separate correctly labeled from incorrectly labeled data. We propose to use one augmentation of the input for sample loss modeling and categorization to create the hypothetical sets $C$ and $I$ or to determine the pseudo label $z$, while utilizing another different augmentation as input to the network $h_\theta$ for purposes of back-propagation. This would require twice the number of forward passes during training for each input. The goal of this is so that we do not adversely affect any loss modeling but also be able to inject more generalization during the learning process.  We provide an example in section \ref{sec:aug-dividemix} for how we can incorporate \textsc{AugDesc} into DivideMix.






\begin{algorithm}[t]
\small
\SetAlgoLined
\hspace*{0em} \textbf{Input}: $\theta^{1}, \theta^{2}$, training batch possibly labeled $x$, possibly unlabeled $u$, dataset labels $y$, gmm probabilities $w$, number of augmentations M, augmentation policies Augment$_1$ and Augment$_2$\\

$x^{desc}  = $ Augment$_2$($x$)\\
$u^{desc}  = $ Augment$_2$($u$)\\

\textbf{for} $m$ = 1 to M \\
\hspace*{1.5em}$x = $ Augment$_1$($x$) \\
\hspace*{1.5em}$u = $ Augment$_1$($u$)\\
\textbf{end}
// co-guessing and sharpening \\
$p = \frac{1}{M}\sum_m p_{model}(x; \theta^{(k)})$\\
$\bar{y} = w y + (1 - w) p$ \\
$\hat{y} = Sharpen(y, T)$ \\
 \begin{tabular}{@{\hspace*{0em}}l@{}}
        $\bar{q} = \frac{1}{2M} \sum_m (p_{model}(\hat{u}; \theta^{(1)}) $\\
        \hspace*{6.5em}$ + p_{model}(\hat{u}; \theta^{(2)}))$
      \end{tabular}\\
$\hat{q} = Sharpen(\bar{q}, T)$ \\ 
// train using a different augmentation \\
$\hat{\mathcal{X}} = \{(x, y) | x \in x^{desc}, y \in \hat{y}\}$ \\ 
$\hat{\mathcal{U}} = \{(u, q) | u \in u^{desc}, q \in \hat{q}\}$ \\
$\mathcal{L}_x, \mathcal{L}_u$ = MixMatch($\hat{\mathcal{X}}, \hat{\mathcal{U}}$) \\
$\mathcal{L} = \mathcal{L}_x + \lambda_u \mathcal{L}_u + \lambda_r \mathcal{L}_{reg}$ \\
$\theta^{(k)} = $SGD($\mathcal{L}, \theta^{(k)})$

 \caption{Batch level training modifications to DivideMix for Augmented Descent. Full implementation provided in the supplemental.}
 \label{algo:dividemix}
\end{algorithm}






\subsection{Augmentation Strategies}
\label{sec:aug-strat-description}

We examine the following strategies for incorporating augmentation into existing algorithms. Figure \ref{fig:strat} presents a conceptual representation for incorporating our augmentation strategy into existing techniques.

\textbf{Raw:} Original image is used without any modifications. 

\textbf{Dataset Expansion:} A dataset is created that is twice the original size of the dataset. This is then fed directly into the model without further augmentation.

\textbf{Runtime Augmentation:} Images are transformed before being fed into network at runtime.

\textbf{Augmented Descent (\textsc{AugDesc}):} Two sets of augmented images are created. One set is used for any loss analysis tasks, while the other is used for gradient descent. The motivation is that we can learn a better representation for each image while not compromising the sample filtering and pseudo-labeling process.

\subsection{Augmentation Policy}

We evaluate three different augmentation policies, classified into "weak" and "strong". Many algorithms make use of the standard random crop and flip for augmentation \cite{lin2017focal}. We call this process \textit{weak augmentation}. We experiment with \textit{strong augmentations} using automatically learned policies from AutoAugment \cite{cubuk2019autoaugment} and RandAugment \cite{cubuk2020randaugment}. AutoAugment and RandAugment both provide a way to apply augmentations without hand-tuning the particular policy. Our strong augmentation policy first applies a random crop and flip, followed by an AutoAugment or RandAugment transformation, and lastly normalization. For dataset expansion and runtime augmentation, we experiment with both weak and strong augmentations. 

We examine three variants of Augmented Descent. \textsc{AugDesc-WW} means we perform loss analysis using a weakly-augmented input, then use this label to train a different weakly augmented version of the same input. Similarly, \textsc{AugDesc-SS} represents strongly-augmented loss analysis, coupled with strongly augmented gradient descent. Finally, \textsc{AugDesc-WS} corresponds to weakly-augmented loss analysis with strongly augmented optimization.

Because AutoAugment is learned on a small subset of the actual data, it is easy to incorporate into existing architectures. We further perform an ablation study using RandAugment to show that our augmentation strategy is agnostic to augmentation policy, as well as the fact that no dataset-specific or pre-trained augmentations are necessary. We use AutoAugment for most of our experiments as it prescribes a pre-trained set of policies, while RandAugment requires tuning that can depend on the networks used as well as the training set size.

\subsection{Application to State of the Art}
\label{sec:aug-dividemix}
While many techniques beyond those above have similar characteristics that we can analyze in a similar manner, we examine this augmentation strategy within the context of the current state-of-the-art DivideMix \cite{li2020dividemix} in this paper. We then extend our augmentation strategy to other techniques and report results in the experiments section.

DivideMix incorporates aspects of warm-up, co-training\cite{jiang2018mentornet, han2018coteaching}, and MixUp \cite{zhang2017mixup}. The original DivideMix algorithm works by first warming up using normal cross-entropy loss with a penalty for confident predictions by adding a negative cross entropy term from Pereyra et al. \cite{pereyra2017regularizing}. Afterwards, for each training epoch, the algorithm first uses a GMM to model the per-sample loss with each of the two networks. Using this and a clean probability threshold, the network then categorizes samples into a labeled set $x$ and an unlabeled set $u$.  Batches are pulled from from each of these two sets and are first augmented. Predictions using the augmented samples are made and a sharpening function is applied to the output \cite{berthelot2019mixmatch} to reduce the entropy of the label distribution. This produces sharpened guesses for the labeled and unlabeled inputs which is used for optimization.

We outline the application of our augmentation strategy in Algorithm \ref{algo:dividemix}. We require two different sets of augmentations: one for the original DivideMix pipeline, and one to augment the original input for training with MixMatch losses. Additional examples of implementation in previous techniques are included in the supplemental.

\section{Experiments}

We first perform evaluations on synthetically generated noise to determine an effective augmentation strategy. We then conduct generalization experiments on real-world datasets, apply our strategies to previous techniques, and experiment with alternative augmentation policies.

\subsection{Experimental Setup}

We perform extensive validation of each augmentation technique on CIFAR-10 and CIFAR-100, two well-known synthetic image classification datasets frequently used for this task. CIFAR-10 contains 10 categories of images and CIFAR-100 contains 100 categories for classification. Each dataset has 50K color images for training and 10K test images of size 32x32. Symmetric and asymmetric noise injection methods \cite{tanaka2018joint, li2019learning} are evaluated. We perform most of the ablation studies within the DivideMix framework as this is the state-of-the-art technique. We then extend the augmentation strategies we found to other techniques.

We use an 18-layer PreAct Resnet \cite{he2016identity} as the network backbone and train it using SGD with a batch size of 128. Some experiments are conducted using a batch size of 64 due to hardware constraints but consistency is maintained in the comparisons. We conduct the experiments using the method outlined in \cite{li2020dividemix} with all the same hyperparameters: a momentum of 0.9, weight decay of 0.0005, and trained for roughly 300 epochs depending on the speed of convergence. The initial learning rate is set to 0.02 and reduced by a factor of 10 after roughly 150 epochs. Warm-up periods where applicable are set to 10 epochs for CIFAR-10 and to 30 epochs for CIFAR-100. We keep the number of augmentations parameter $M = 2$ fixed for a fair comparison.

\subsection{Comparison of Augmentation Strategies}
\label{sec:aug-strat}

\begin{table}[t]
\centering
\scalebox{0.9}{
    \begin{tabular}{ll|rr|rr}
                 &      & \multicolumn{2}{c|}{CIFAR-10} & \multicolumn{2}{c}{CIFAR-100} \\
    Method/Noise             &      & 20\%         & 90\%         & 20\%          & 90\%         \\ \hline
    Raw       & Best &       85.94       &      27.58        &       52.24        &     7.99         \\
                 & Last &    83.23          &     23.92         &     39.18          &    2.98          \\\hline
    Expansion-W    & Best &     90.86         &     31.22         &    57.11           &      7.30        \\
                 & Last &        89.95      &      10.00        &  53.29         &      2.23     \\\hline
    Expansion-S    & Best &      90.56        &     35.10         &     55.15          &     7.54         \\
                 & Last &     89.51        &       34.23       &      54.37         &     3.24         \\\hline
    Runtime-W \cite{li2020dividemix}   & Best &   96.10           & 76.00           & 77.30              & 31.50         \\
                 & Last &   95.70           & 75.40         &  76.90             & 31.00           \\\hline
    Runtime-S & Best &       \textbf{96.54}      & 70.47        &      \textbf{79.89}         & 40.52        \\
                 & Last &     \textbf{96.33}         & 70.22        &    79.40           & 40.34        \\\hline
    AugDesc-WW   & Best &    96.27          & 36.05        &       78.90        & 30.33        \\
                 & Last &    96.08          & 23.50         &     78.44          & 29.88        \\\hline
    AugDesc-SS   & Best & 96.47        & 81.77        &    79.79           & 38.85        \\
                 & Last & 96.19         & 81.54     &         \textbf{79.51}          & 38.55        \\\hline
    AugDesc-WS   & Best & 96.33        & \textbf{91.88}        & 79.50          & \textbf{41.20}         \\
                 & Last & 96.17        & \textbf{91.76}        & 79.22         & \textbf{40.90}        
    \end{tabular}
}
\caption{Performance differences for each augmentation strategy. The best performance in each category is highlighted in bold. Removing all augmentation is highly detrimental to performance, while more augmentation seemingly improves performance. However, too much augmentation is also detrimental to performance (AugDesc-SS). Strategically adding augmentation by exploiting the loss properties (AugDesc-WS) yields the best results in general.}
\label{tab:aug-strat}
\end{table}

We examine the performance of each proposed augmentation strategy outlined in Section \ref{sec:aug-strat-description} using DivideMix as our baseline model. We investigate the performance impact on lower label noise (20\%) and very high label noise (90\%) for some performance bounds. We report results in Table \ref{tab:aug-strat}. 

As shown in the table, there is a large effect on algorithm performance based on how augmentations are included. While in some aspects this is unsurprising, what is surprising is the huge effect augmentation can have with regards to higher noise datasets. In the best case, we see \textsc{AugDesc-WS} at 90\% noise achieve results on CIFAR-10 close to accuracies reported on augmentation techniques with 20\% label noise. For CIFAR-100, we also witness a large effect with higher noise rates but it remains a challenging benchmark for noisy datasets. Overall, we find that AugDesc-WS achieves the strongest result across the board. 

It should be noted that a vast number of image-based machine learning algorithms incorporate some level of weak augmentation (flip, crop, and normalization) during training time. For completeness, we retrospectively examine the effect of removing these augmentations to tease out the effect of augmentation, i.e. the raw input method. We see that including some very small amount of augmentation is hugely beneficial, particularly evident when examining the transition from raw to weak augmentation at runtime.


\subsection{Effect of Augmentation During Warm-up}
\label{sec:warmup}

\begin{figure*}[t]
\includegraphics[width=\textwidth]{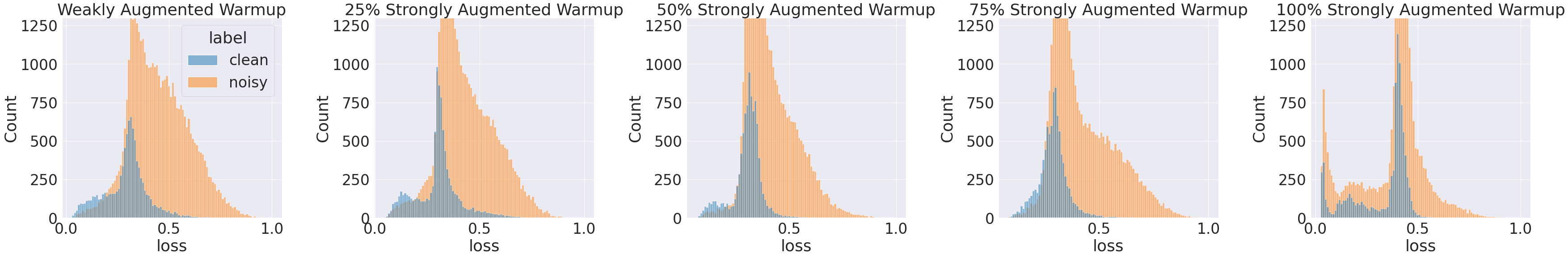}
\caption{Effect of augmentation strength on the distribution of normalized loss for noisy versus clean segments of the dataset during warm-up for 90\% label noise. Too much augmentation can cause samples in the clean dataset to be have higher loss, causing lower loss in samples from the noisy dataset.}
\label{fig:aug-chance}
\end{figure*}

\begin{table*}[t]
\centering
\scalebox{0.9}{
\begin{tabular}{llrrrr|c|rrrr}
                  &       & \multicolumn{5}{c|}{\textbf{CIFAR-10}}              & \multicolumn{4}{c}{\textbf{CIFAR-100}}  \\ \cline{3-11} 
  Model                     & Noise & 20\%  & 50\%  & 80\%  & 90\%  & 40\% Asym & 20\%  & 50\%  & 80\%  & 90\%  \\ \hline
                       
\multirow{2}{*}{DivideMix (baseline) \cite{li2020dividemix} }    & Best  & 96.1  & 94.6  & 92.3  & 76.0    & 93.4      & 77.3  & 74.6  & 60.2  & 31.5  \\
                       & Last  & 95.7  & 94.4  & 92.9  & 75.4 & 92.1  & 76.9  & 74.2  & 59.6  & 31.0    \\\hline
\multirow{2}{*}{DM-AugDesc-WS-SAW} & Best       & \textbf{96.3} & \textbf{95.6} & 93.7  & 35.3 & 94.4     &   \textbf{79.6}    &   \textbf{77.6}    &   61.8    & 17.3 \\
                       & Last                   & \textbf{96.2} & \textbf{95.4} &  \textbf{93.6} & 10.0    & 94.1     &  \textbf{79.5}     &    \textbf{77.5}   &   61.6    & 15.1 \\\hline
\multirow{2}{*}{DM-AugDesc-WS-WAW} & Best       & \textbf{96.3} & 95.4 & \textbf{93.8} & \textbf{91.9} & \textbf{94.6}     & 79.5  & 77.2 & \textbf{66.4} & \textbf{41.2} \\
                   & Last                       & \textbf{96.2} & 95.1 & \textbf{93.6} & \textbf{91.8} & \textbf{94.3}     & 79.2 & 77.0 & \textbf{66.1} & \textbf{40.9}  
\end{tabular}
}
\caption{Application of strong versus weak augmentation during the warm-up period of DivideMix, in comparison to the baseline model. WAW signifies weakly augmented warm-up, SAW represents strongly augmented warm-up. Weak warm-up appears to benefit datasets with higher noise while strong warm-up benefits datasets with lower noise.}
\label{tab:warm-up}
\end{table*}

LNL algorithms generally rely on fact that clean samples are fit before noisy ones. To take advantage of such a property, many algorithms create scheduled learning or tune the loss function, explicitly designating warm-up period to exploit the label noise learning property \cite{arazo2019unsupervised, li2020dividemix, yu2019does}. We test the effect of introducing augmentation before and after this period by comparing the performance of models injected with augmentations from the first epoch and models trained with augmentations after the designated warm-up period.

We report performance metrics in Table \ref{tab:warm-up} for various noise levels. We find that injecting strong augmentations during the warm-up period in low noise datasets benefit performance, but is detrimental when the dataset becomes increasingly noisy. This is particularly evident when examining the 90\% noise rate.  Conversely, weakly augmented warm-up greatly increases performance at higher noise levels.

To better understand why this is, we perform an experiment by stochastically applying strong augmentation to each batch with increasing chance to observe its distribution at epoch 20. Figure \ref{fig:aug-chance} shows the loss distribution for samples in the training set associated with the clean versus the noisy dataset. We find that applying too much augmentation too soon can encourage lower noise data to have too high of a loss and noisy data to have lower loss.


\subsection{Synthetic Dataset Summary Results}
\label{sec:synthetic}

\begin{table*}[t]
\centering
\scalebox{0.9}{
\begin{tabular}{llrrrr|rrrrr}
  &  &  \multicolumn{4}{c|}{\textbf{CIFAR-10}}  &  \multicolumn{4}{c}{\textbf{CIFAR-100}}  \\  \cline{3-10}  
  Model  &  Noise  &  20\%  &  50\%  &  80\%  &  90\%  &  20\%  &  50\%  &  80\%  &  90\%  \\  \hline
  
\multirow{2}{*}{Cross-Entropy}  &  Best  &  86.8  &  79.4  &  62.9  &  42.7  &  62.0  &  46.7  &  19.9  &  10.1  \\
  &  Last  &  82.7  &  57.9  &  26.1  &  16.8  &  61.8  &  37.3  &  8.8  &  3.5  \\\hline
\multirow{2}{*}{Reed et. al. \cite{reed2014training}}  &  Best  &  86.8  &  79.8  &  63.3  &  42.9  &  62.1  &  46.6  &  19.9  &  10.2  \\
  &  Last  &  82.9  &  58.4  &  26.8  &  17.0  &  62.0  &  37.9  &  8.9  &  3.8  \\\hline
\multirow{2}{*}{Yu et al. \cite{yu2019does}}  &  Best  &  89.5  &  85.7  &  67.4  &  47.9  &  65.6  &  51.8  &  27.9  &  13.7  \\
  &  Last  &  88.2  &  84.1  &  45.5  &  30.1  &  64.1  &  45.3  &  15.5  &  8.8  \\\hline
\multirow{2}{*}{Zhang et al. \cite{zhang2017mixup}}  &  Best  &  95.6  &  87.1  &  71.6  &  52.2  &  67.8  &  57.3  &  30.8  &  14.6  \\
  &  Last  &  92.3  &  77.6  &  46.7  &  43.9  &  66.0  &  46.6  &  17.6  &  8.1  \\\hline
\multirow{2}{*}{Yi \& Wu \cite{yi2019probabilistic}}  &  Best  &  92.4  &  89.1  &  77.5  &  58.9  &  69.4  &  57.5  &  31.1  &  15.3  \\
  &  Last  &  92.0  &  88.7  &  76.5  &  58.2  &  68.1  &  56.4  &  20.7  &  8.8  \\\hline
\multirow{2}{*}{Li et al. \cite{li2019learning}}  &  Best  &  92.9  &  89.3  &  77.4  &  58.7  &  68.5  &  59.2  &  42.4  &  19.5  \\
  &  Last  &  92.0  &  88.8  &  76.1  &  58.3  &  67.7  &  58.0  &  40.1  &  14.3  \\\hline
\multirow{2}{*}{Arazo et al. \cite{arazo2019unsupervised}}  &  Best  &  94.0  &  92.0  &  86.8  &  69.1  &  73.9  &  66.1  &  48.2  &  24.3  \\
  &  Last  &  93.8  &  91.9  &  86.6  &  68.7  &  73.4  &  65.4  &  47.6  &  20.5  \\\hline
\multirow{2}{*}{Li et al. \cite{li2020dividemix}}  &  Best  &  96.1  &  94.6  &  92.9  &  76.0  &  77.3  &  74.6  &  60.2  &  31.5  \\
  &  Last  &  95.7  &  94.4  &  92.3  &  75.4  &  76.9  &  74.2  &  59.6  &  31.0  \\\hline
\multirow{2}{*}{DM-AugDesc-WS-SAW}  &  Best  &  \textbf{96.3}&  \textbf{95.6}  &  93.7  &  35.3  &  \textbf{79.6}  &  \textbf{77.6}  &  61.8  &  17.3  \\
  &  Last  &  \textbf{96.2}  &\textbf{95.4}&  \textbf{93.6}  &  10.0  &  \textbf{79.5}  &  \textbf{77.5}  &  61.6  &  15.1 \\\hline
\multirow{2}{*}{DM-AugDesc-WS-WAW}  &  Best  &  \textbf{96.3}  &  95.4  &  \textbf{93.8}  &  \textbf{91.9}  &  79.5  &  77.2  &  \textbf{66.4}  &  \textbf{41.2}  \\
  &  Last  &  \textbf{96.2}  &  95.1  &  \textbf{93.6}  &  \textbf{91.8}  &  79.2  &  77.0  &  \textbf{66.1}  &  \textbf{40.9} 
\end{tabular}
}

\caption{Performance comparison when incorporating our best augmentation strategy into the current state-of-the-art. Our augmentation strategy improves performance at every noise level. Results for previous techniques were directly copied from their respective papers.}
\label{tab:summary}
\end{table*}

We report the summary results in Table \ref{tab:summary}. The results show that augmenting the state-of-the-art algorithm using our best augmentation strategy increases accuracy across all noise levels. In particular, the improvement for extremely noisy datasets (90\%) is very large, and approaches the best performance of lower noise datasets and represents an error reduction of 65\%. For comparison, we achieve 91\% accuracy for 90\% symmetric noise on the CIFAR-10 dataset while the previous state of the art achieves 96.1\% on only 20\% label noise. Furthermore, we achieve an over 15\% improvement in accuracy over previous state of the art for CIFAR-10 at 90\% label noise.

\subsection{Clothing1M Performance}
\label{sec:clo1m}

\begin{table}[t]
\centering

\scalebox{0.9}{
\begin{tabular}{l|c}
Method & Test Accuracy \\ \hline
Cross Entropy & 69.21 \\
M-correction \cite{arazo2019unsupervised} & 71.00 \\
Joint Optimization \cite{tanaka2018joint} & 72.16 \\
MetaCleaner \cite{zhang2019metacleaner} & 72.50\\
MLNT \cite{li2019learning}     &    73.47           \\
PENCIL \cite{yi2019probabilistic} & 73.49 \\
DivideMix \cite{li2020dividemix}       &        74.76 \\
ELR+ \cite{liu2020early} & 74.81 \\ \hline
DM-AugDesc-WS-WAW (ours) & 74.72 \\
\textbf{DM-AugDesc-WS-SAW (ours)}       &    \textbf{75.11}        
\end{tabular}
}
\caption{Comparison against state-of-the-art methods for accuracy on the Clothing1M dataset.}
\label{tab:clo1m}
\end{table}

Clothing1M \cite{xiao2015learning} is a large-scale real-world dataset containing 1 million images obtained from online shopping websites. Labels are generated by extracting tags from the surrounding texts and keywords, and are thus very noisy. A ResNet-50 with pre-trained ImageNet weights are used following the work of \cite{li2019learning}. We applied the pre-trained ImageNet AutoAugment augmentation policy for this task. 

We report results in table \ref{tab:clo1m}. Our augmentation strategy obtained state-of-the-art performance when utilizing a strongly augmented warm-up cycle. In addition to obtaining competitive results, this further indicates that the noise level is likely to be below 80\% based on our previous experiments, as strong warm-up improves accuracy. This is in concordance with the estimates of the noise level of Clothing1M, said to be approximately 61.54\% \cite{xiao2015learning}.

\subsection{Automatic Augmentation Policies}

\begin{table}[t]
\centering
\scalebox{0.9}{
\begin{tabular}{ll|rr|rr}
             &      & \multicolumn{2}{c|}{CIFAR-10} & \multicolumn{2}{c}{CIFAR-100} \\
Method/Noise             &      & 20\%         & 90\%         & 20\%          & 90\%         \\ \hline

Baseline \cite{li2020dividemix}   & Best &   96.1           & 76.0           & 77.3              & 31.5         \\
             & Last &   95.7           & 75.4         &  76.9             & 31.0           \\\hline
             
AutoAugment    & Best &     \textbf{96.3}        &       \textbf{91.9}       &       \textbf{79.5}        &      \textbf{41.2}        \\
             & Last &       \textbf{96.2}       &            \textbf{91.8}  &      \textbf{79.2}         &         \textbf{40.9}     \\\hline
RandAugment    & Best &   96.1           &       89.6       &      78.1         &     36.8         \\
             & Last &       96.0       &       89.4       &         77.8      &     36.7         \\     
\end{tabular}
}
\caption{Comparison of different automated augmentation policy algorithms. We compare performance of each policy using the AugDesc-WS approach. Adjusting the augmentation policy has minimal effect but still handily outperforms the runtime augmentation used in the baseline. The improved performance is still large with a noise ratio of 90\%.}
\label{tab:aug-policy}
\end{table}

In our evaluation benchmarks, we primarily used AutoAugment pre-trained policies. These policies are trained on a small subset of the original dataset with regards to CIFAR-10 and CIFAR-100 (5000 samples). We do this due to the simplistic nature of integrating pre-trained AutoAugment policies. For completeness, we evaluate whether we can achieve similar performance with an untrained set of augmentations, as theoretically we could then tune policies based on validation accuracy. To do this, we examine whether we can achieve performance on-par with AutoAugment using RandAugment \cite{cubuk2020randaugment}, which can be tuned by adjusting 2 parameters. For these experiments, we used $N = 1$ and $M = 6$ for RandAugment hyperparameters.

We report results in Table \ref{tab:aug-policy}. As shown in the table, RandAugment can achieve performance on-par with AutoAugment with minimal tuning and demonstrates the validity of our approach. Furthermore, since we were able to outperform the state-of-the-art on Clothing1M while using a pre-trained ImageNet AutoAugment policy for the task, optimizing an AutoAugment policy on Clothing1M could potentially yield better results.

\subsection{Generalization to Previous Techniques}
\label{sec:generalization}
Based on our evaluations, we find that a weakly augmented warm-up period followed by the application of strong augmentation works best. Furthermore, it is beneficial to perform the loss analysis process on a weakly augmented input, then forwarding a strongly augmented input through the network for training. We apply our most effective augmentation strategy to previous techniques to evaluate generalizability of our approach. 

We choose to compare to Cross-Entropy, Co-Teaching+\cite{yu2019does}, M-DYR-H \cite{arazo2019unsupervised}, and DivideMix \cite{li2020dividemix} due to the range of techniques these algorithms employ. Co-Teaching+ uses two networks and thresholding to exploit the memorization effect and is an updated work based on the popular Co-Teaching \cite{han2018coteaching} technique. M-DYR-H uses mixture models to fit the loss to previous epochs to weight the models predictions using a single network. DivideMix is the current state-of-the-art which combines these and brings in a semi-supervised learning framework.

All source code for each evaluated technique was available publicly published by the original authors. We follow the hyperparameters and models outlined in the original published paper and apply no tuning of our own. This demonstrates the ease at which augmentations can be incorporated without delicate tuning of hyperparameters, highlighting the generalizability of our approach. We detail the exact algorithm modifications for inserting augmentations in the supplemental of this paper. We perform the evaluation on a lower noise setting (20\%) as many previous techniques did not perform well at high noise levels. Table \ref{tab:generalization} shows the performance of our evaluation.

For vanilla cross-entropy, we used \textsc{Runtime-S} since as there is no warm-up period. For other techniques, we applied the \textsc{AugDesc-WS-WAW} strategy. We evaluated our augmentation strategy on these algorithms as they cover a range of general approaches to learning with label noise. Some differences in performance are larger than expected due to the specific implementation of network architecture and synthetic noise generation techniques. We attempted strongly augmented warm-up for Co-teaching and found that there was a very large detrimental impact to performance. This agrees with our earlier observation that too much augmentation during the warm-up period can be detrimental. In particular, it appears to have a strong impact on the way noisy and clean data converge during the warm-up period, which these algorithms typically rely on.

The \textsc{AugDesc-WS-WAW} strategy and even augmentation in general benefits performance in multiple categories (Table \ref{tab:generalization}). As the experiments conducted were with no tuning of hyperparameters, we expect that further improvements can be seen when tuning with augmentation in mind due to the ways in which these algorithms exploit the loss distributions. Additionally, we see that across the board, the average performance of the last few epochs with augmentation is better than performance without. This indicates that using our augmentation strategy aids in learning a better distribution.

\begin{table}[t!]
\centering

\scalebox{0.9}{
\begin{tabular}{llrr|rr}
                               &      & \multicolumn{2}{c|}{\textbf{CIFAR-10}}                    & \multicolumn{2}{c}{\textbf{CIFAR-100}}                                                       \\ \cline{3-6}
                               &      & \textbf{Base}               & \textbf{Aug}                & \textbf{Base}              & \textbf{Aug}                \\ \hline
\multirow{2}{*}{Cross Entropy} & Best & 86.8                 &    \textbf{89.9}           & 60.2                   &     \textbf{61.2}                    \\
                               & Last & 82.7                 &   \textbf{85.1}               & 59.9                &  \textbf{ 60.4}                      \\ \hline
\multirow{2}{*}{Co-Teaching+ \cite{yu2019does}}   & Best & 59.3                & \textbf{60.6}                 & \textbf{26.2}                & 25.6                       \\
                               & Last & 55.9                & \textbf{57.4}                 & 23.0                & \textbf{23.7}                      \\ \hline
\multirow{2}{*}{M-DYR-H \cite{arazo2019unsupervised}}  & Best & \textbf{94.0}                   &    93.9                       & 68.2                 &    \textbf{73.0}                          \\
                               & Last & 93.8                 &    \textbf{93.9}                         & 67.5                 &         \textbf{72.7}                       \\ \hline
\multirow{2}{*}{DivideMix}     & Best & 96.1                 & \textbf{96.3}                  & 77.3                 &      \textbf{79.5}                \\
                               & Last & 95.7                 & \textbf{96.2}              & 76.9                 &      \textbf{79.2}                      
\end{tabular}
}
\caption{Performance benefits when applying our augmentation strategy to previous techniques at 20\% noise level. Baseline and augmented accuracy scores are reported.}
\label{tab:generalization}
\end{table}

\section{Conclusion}
In this paper, we propose and examine the effect of various augmentation strategies within the domain of learning with label noise. We find that it is advantageous to add additional augmentation, particularly for higher noise ratios. Furthermore, copious amounts of augmentation during warm-up periods should be avoided if the noise rate is high, as this can have detrimental effects on the property that neural networks fit clean data before noisy data \cite{arpit2017closer}. We performed extensive studies and found that the \textsc{AugDesc-WS} strategy is capable of producing improvements across all noise levels and in multiple datasets. We further show its generalization capabilities by applying it to previous techniques with demonstrated success. This is additional evidence for how using two separate pools of augmentation operations for two separate tasks in these machine learning algorithms can be beneficial. This idea has previously been demonstrated to be effective in SSL settings \cite{sohn2020fixmatch}, and we now show this for LNL settings.

In summary, we examined where it is advantageous to incorporate varying degrees of augmentation, and were able to demonstrate a strategy to advance the state-of-the-art as well as improve the performance of previous techniques. We hope the insights regarding the strength and amount of augmentation will be beneficial for future applications of augmentation when developing LNL algorithms.

\section{Acknowledgements}
This work was supported in part by ONR awards N00014-19-1-2553 and N00174-19-1-0024, as well as NSF awards IIS-1911230 and IIS-1845587. We would also like to thank Dr. Lina Kim and all those involved with the UCSB RMP program.

{\small
\bibliographystyle{ieee_fullname}
\bibliography{main}

\begin{thebibliography}{10}\itemsep=-1pt

\bibitem{arazo2019unsupervised}
Eric Arazo, Diego Ortego, Paul Albert, Noel~E O'Connor, and Kevin McGuinness.
\newblock Unsupervised label noise modeling and loss correction.
\newblock {\em arXiv preprint arXiv:1904.11238}, 2019.

\bibitem{arpit2017closer}
Devansh Arpit, Stanisław Jastrzębski, Nicolas Ballas, David Krueger, Emmanuel
  Bengio, Maxinder~S. Kanwal, Tegan Maharaj, Asja Fischer, Aaron Courville,
  Yoshua Bengio, and Simon Lacoste-Julien.
\newblock A closer look at memorization in deep networks.
\newblock In {\em ICML}, 2017.

\bibitem{berthelot2020remix}
David Berthelot, Nicholas Carlini, Ekin~D. Cubuk, Alex Kurakin, Kihyuk Sohn,
  Han Zhang, and Colin Raffel.
\newblock Remixmatch: Semi-supervised learning with distribution matching and
  augmentation anchoring.
\newblock In {\em ICLR}, 2020.

\bibitem{berthelot2019mixmatch}
David Berthelot, Nicholas Carlini, Ian Goodfellow, Nicolas Papernot, Avital
  Oliver, and Colin~A Raffel.
\newblock Mixmatch: A holistic approach to semi-supervised learning.
\newblock In {\em Advances in Neural Information Processing Systems}, pages
  5049--5059, 2019.

\bibitem{chen2019understanding}
Pengfei Chen, Ben~Ben Liao, Guangyong Chen, and Shengyu Zhang.
\newblock Understanding and utilizing deep neural networks trained with noisy
  labels.
\newblock In {\em ICML}, 2019.

\bibitem{cubuk2019autoaugment}
Ekin~D Cubuk, Barret Zoph, Dandelion Mane, Vijay Vasudevan, and Quoc~V Le.
\newblock Autoaugment: Learning augmentation strategies from data.
\newblock In {\em Proceedings of the IEEE conference on computer vision and
  pattern recognition}, pages 113--123, 2019.

\bibitem{cubuk2020randaugment}
Ekin~D Cubuk, Barret Zoph, Jonathon Shlens, and Quoc~V Le.
\newblock Randaugment: Practical automated data augmentation with a reduced
  search space.
\newblock In {\em Proceedings of the IEEE/CVF Conference on Computer Vision and
  Pattern Recognition Workshops}, pages 702--703, 2020.

\bibitem{devries2017improved}
Terrance DeVries and Graham~W Taylor.
\newblock Improved regularization of convolutional neural networks with cutout.
\newblock {\em arXiv preprint arXiv:1708.04552}, 2017.

\bibitem{goldberger2017smodel}
Jacob Goldberger and Ehud Ben-Reuven.
\newblock Training deep neural-networks using a noise adaptation layer.
\newblock In {\em ICLR}, 2017.

\bibitem{han2018coteaching}
Bo Han, Quanming Yao, Xingrui Yu, Gang Niu, Miao Xu, Weihua Hu, Ivor Tsang, and
  Masashi Sugiyama.
\newblock Co-teaching: Robust training of deep neural networks with extremely
  noisy labels.
\newblock In {\em NeurIPS}, pages 8535--8545, 2018.

\bibitem{he2016identity}
Kaiming He, Xiangyu Zhang, Shaoqing Ren, and Jian Sun.
\newblock Identity mappings in deep residual networks.
\newblock In {\em European conference on computer vision}, pages 630--645.
  Springer, 2016.

\bibitem{hendrycks2019augmix}
Dan Hendrycks, Norman Mu, Ekin~D Cubuk, Barret Zoph, Justin Gilmer, and Balaji
  Lakshminarayanan.
\newblock Augmix: A simple data processing method to improve robustness and
  uncertainty.
\newblock {\em arXiv preprint arXiv:1912.02781}, 2019.

\bibitem{jiang2018mentornet}
Lu Jiang, Zhengyuan Zhou, Thomas Leung, Li-Jia Li, and Li Fei-Fei.
\newblock Mentornet: Learning data-driven curriculum for very deep neural
  networks on corrupted labels.
\newblock In {\em ICML}, 2018.

\bibitem{li2020dividemix}
Junnan Li, Richard Socher, and Steven~CH Hoi.
\newblock Dividemix: Learning with noisy labels as semi-supervised learning.
\newblock {\em arXiv preprint arXiv:2002.07394}, 2020.

\bibitem{li2019learning}
Junnan Li, Yongkang Wong, Qi Zhao, and Mohan~S Kankanhalli.
\newblock Learning to learn from noisy labeled data.
\newblock In {\em Proceedings of the IEEE Conference on Computer Vision and
  Pattern Recognition}, pages 5051--5059, 2019.

\bibitem{lin2017focal}
Tsung-Yi Lin, Priya Goyal, Ross Girshick, Kaiming He, and Piotr Doll{\'a}r.
\newblock Focal loss for dense object detection.
\newblock In {\em Proceedings of the IEEE international conference on computer
  vision}, pages 2980--2988, 2017.

\bibitem{liu2020early}
Sheng Liu, Jonathan Niles-Weed, Narges Razavian, and Carlos Fernandez-Granda.
\newblock Early-learning regularization prevents memorization of noisy labels.
\newblock {\em arXiv preprint arXiv:2007.00151}, 2020.

\bibitem{luo2018smooth}
Yucen Luo, Jun Zhu, Mengxi Li, Yong Ren, and Bo Zhang.
\newblock Smooth neighbors on teacher graphs for semi-supervised learning.
\newblock In {\em Proceedings of the IEEE conference on computer vision and
  pattern recognition}, pages 8896--8905, 2018.

\bibitem{ma2018dimensionalitydriven}
Xingjun Ma, Yisen Wang, Michael~E. Houle, Shuo Zhou, Sarah~M. Erfani, Shu-Tao
  Xia, Sudanthi Wijewickrema, and James Bailey.
\newblock Dimensionality-driven learning with noisy labels.
\newblock In {\em ICML}, 2018.

\bibitem{malach2017decoupling}
Eran Malach and Shai Shalev-Shwartz.
\newblock Decoupling “when to update” from “how to update”.
\newblock In {\em NIPS}, 2017.

\bibitem{natarajan2013learning}
Nagarajan Natarajan, Inderjit~S Dhillon, Pradeep~K Ravikumar, and Ambuj Tewari.
\newblock Learning with noisy labels.
\newblock In {\em Advances in neural information processing systems}, pages
  1196--1204, 2013.

\bibitem{patrini2017forward}
Giorgio Patrini, Alessandro Rozza, Aditya~Krishna Menon, Richard Nock, and
  Lizhen Qu.
\newblock Making deep neural networks robust to label noise: a loss correction
  approach.
\newblock In {\em CVPR}, 2017.

\bibitem{pereyra2017regularizing}
Gabriel Pereyra, George Tucker, Jan Chorowski, {\L}ukasz Kaiser, and Geoffrey
  Hinton.
\newblock Regularizing neural networks by penalizing confident output
  distributions.
\newblock {\em arXiv preprint arXiv:1701.06548}, 2017.

\bibitem{reed2014training}
Scott Reed, Honglak Lee, Dragomir Anguelov, Christian Szegedy, Dumitru Erhan,
  and Andrew Rabinovich.
\newblock Training deep neural networks on noisy labels with bootstrapping.
\newblock {\em arXiv preprint arXiv:1412.6596}, 2014.

\bibitem{Reed2014bootstrap}
Scott~E. Reed, Honglak Lee, Dragomir Anguelov, Christian Szegedy, Dumitru
  Erhan, and Andrew Rabinovich.
\newblock Training deep neural networks on noisy labels with bootstrapping.
\newblock In {\em ICLR}, 2015.

\bibitem{sajjadi2016consistency}
Mehdi Sajjadi, Mehran Javanmardi, and Tolga Tasdizen.
\newblock Regularization with stochastic transformations and perturbations for
  deep semi-supervised learning.
\newblock In {\em Advances in Neural Information Processing Systems}, 2016.

\bibitem{sohn2020fixmatch}
Kihyuk Sohn, David Berthelot, Chun-Liang Li, Zizhao Zhang, Nicholas Carlini,
  Ekin~D Cubuk, Alex Kurakin, Han Zhang, and Colin Raffel.
\newblock Fixmatch: Simplifying semi-supervised learning with consistency and
  confidence.
\newblock {\em arXiv preprint arXiv:2001.07685}, 2020.

\bibitem{sohn2020simple}
Kihyuk Sohn, Zizhao Zhang, Chun-Liang Li, Han Zhang, Chen-Yu Lee, and Tomas
  Pfister.
\newblock A simple semi-supervised learning framework for object detection.
\newblock {\em arXiv preprint arXiv:2005.04757}, 2020.

\bibitem{tanaka2018joint}
Daiki Tanaka, Daiki Ikami, Toshihiko Yamasaki, and Kiyoharu Aizawa.
\newblock Joint optimization framework for learning with noisy labels.
\newblock In {\em Proceedings of the IEEE Conference on Computer Vision and
  Pattern Recognition}, pages 5552--5560, 2018.

\bibitem{xiao2015learning}
Tong Xiao, Tian Xia, Yi Yang, Chang Huang, and Xiaogang Wang.
\newblock Learning from massive noisy labeled data for image classification.
\newblock In {\em Proceedings of the IEEE conference on computer vision and
  pattern recognition}, pages 2691--2699, 2015.

\bibitem{xie2020uda}
Qizhe Xie, Zihang Dai, Eduard Hovy, Minh-Thang Luong, and Quoc~V. Le1.
\newblock Unsupervised data augmentation for consistency training.
\newblock In {\em NeurIPS}, 2020.

\bibitem{xie2020self}
Qizhe Xie, Minh-Thang Luong, Eduard Hovy, and Quoc~V Le.
\newblock Self-training with noisy student improves imagenet classification.
\newblock In {\em Proceedings of the IEEE/CVF Conference on Computer Vision and
  Pattern Recognition}, pages 10687--10698, 2020.

\bibitem{yi2019probabilistic}
Kun Yi and Jianxin Wu.
\newblock Probabilistic end-to-end noise correction for learning with noisy
  labels.
\newblock In {\em Proceedings of the IEEE Conference on Computer Vision and
  Pattern Recognition}, pages 7017--7025, 2019.

\bibitem{yu2019does}
Xingrui Yu, Bo Han, Jiangchao Yao, Gang Niu, Ivor~W Tsang, and Masashi
  Sugiyama.
\newblock How does disagreement help generalization against label corruption?
\newblock {\em arXiv preprint arXiv:1901.04215}, 2019.

\bibitem{zhang2017mixup}
Hongyi Zhang, Moustapha Cisse, Yann~N Dauphin, and David Lopez-Paz.
\newblock mixup: Beyond empirical risk minimization.
\newblock {\em arXiv preprint arXiv:1710.09412}, 2017.

\bibitem{zhang2019metacleaner}
Weihe Zhang, Yali Wang, and Yu Qiao.
\newblock Metacleaner: Learning to hallucinate clean representations for
  noisy-labeled visual recognition.
\newblock In {\em Proceedings of the IEEE Conference on Computer Vision and
  Pattern Recognition}, pages 7373--7382, 2019.

\end{thebibliography}
}

\clearpage
\appendix
\section{Example Augmented Algorithms}
We detail the augmented examples from the generalization section here. We provide the modified pseudocode based on published papers and publically available code for the evaluated algorithms Co-Teaching+ \cite{yu2019does}, M-DYR-H \cite{arazo2019unsupervised}, and DivideMix \cite{li2020dividemix}. We bold the region where augmentation is inserted. No hyperparamters were changed in any of the experiments. All models used are the same as those used in the originally published papers.

\subsection{Augmented CoTeaching+}
We provde the full implementation of the Co-Teaching+ \cite{yu2019does} algorithm with our augmentations below. Co-Teaching+ is similar to the original Co-Teaching \cite{han2018coteaching}, but adds a different disagreement component when the two network predictions are not similar. If the predictions are similar, training is conducted on the lower loss samples. We find that adding strong augmentation to this part of the training improves performance. Co-teaching uses a threshold $R(e)$ instead of a mixture model fitting to take advantage of the memorization effect. The two network setup that it uses is an effective technique in existing algorithms.

\begin{algorithm}[h]\small
{\bfseries Input} $\theta^{(1)}$ and $\theta^{(2)}$, training dataset $(\mathcal{X},\mathcal{Y})$, learning rate $\eta$, fixed $\tau$, epoch $T_{k}$ and $T_{\max}$, strong augmentation function \textbf{Augment}.\\

$\theta = \mathrm{WarmUp}(\mathcal{X}, \mathcal{Y}, \theta$) \\

\While{ $e < T_{\max}$}
{
	\For{$b = 1$ \KwTo $B$}
	{	
		{\bfseries Select} prediction disagreement $\bar{x_b}'$ from batch $x_b$; \\
		
		\If {$|\bar{x_b}'| > 0$}{
		
    		 $\bar{x_b}^{'(1)} = \arg\min_{\bar{x_b}':|\bar{x_b}'|\ge \lambda(e)|\bar{x_b}'|}\ell(\bar{x_b}';\theta^{(1)})$; \hfill
    		
    		 $\bar{x_b}^{'(2)} = \arg\min_{\bar{x_b}':|\bar{x_b}'|\ge \lambda(e)|\bar{x_b}'|}\ell(\bar{x_b}';\theta^{(2)})$; \hfill
    
    		$\theta^{(1)} = \theta^{(1)} - \eta\nabla \ell(\bar{x_b}^{'(2)};\theta^{(1)})$; \hfill
    		
    		$\theta^{(2)} = \theta^{(2)} - \eta\nabla \ell(\bar{x_b}^{'(1)};\theta^{(2)})$; \hfill
    		
		}
		
		\Else{
    		 $x_b^{(1)} = \arg\min_{x_b':|x_b'|\ge R(e)|x_b|}\ell(x_b; \theta^{(1)})$; \hfill 
    		
    		 $x_b^{(2)} = \arg\min_{x_b':|x_b'|\ge R(e)|x_b|}\ell(x_b; \theta^{(2)})$; \hfill 
    		
    		 $\theta^{(1)} = \theta^{(1)} - \eta\nabla \ell(\textbf{Augment}(x_b^{(1)}); \theta^{(1)})$; \hfill
    		
    		 $\theta^{(2)} = \theta^{(2)} - \eta\nabla \ell(\textbf{Augment}(x_b^{(2)}); \theta^{(2)})$; \hfill
		}
		
	}

	{\bfseries Update} $R(e) = 1 - \min\left\lbrace  \frac{e}{T_k} \tau, \tau \right\rbrace $;
	
}

{\bfseries Output $\theta^{(1)}$ and $\theta^{(2)}$.}
\caption{Augmented Co-Teaching+}
\label{alg:Co-teaching}
\end{algorithm}

\subsection{Augmenting M-DYR-H}

We provide the full implementation of M-DYR-H algorithm from \cite{arazo2019unsupervised}. M-DYR-H warmup, and uses mixup training on input batches to obtain strong results. The loss is weighted using a BMM that is fit to the loss from previous epochs. During warmup, we leave the existing weak-augmentations in place. For the pseudolabel prediction $z_b$ as well as the BMM modelling $W$, we use weak augmentations. We insert strong augmentations during the mixup process which is independent of what the network uses to model the losses. We find that this can improve performance. 

\begin{algorithm}[h]

	\DontPrintSemicolon
	\small
	\textbf{Input:} $\theta$, training dataset $(\mathcal{X},\mathcal{Y})$,  $\mathrm{Beta}$ distribution parameter $\alpha$ for $\mathrm{mixup}$, strong augmentation function \textbf{Augment}. \\

	$\theta=\mathrm{WarmUp}(\mathcal{X},\mathcal{Y},\theta)$ \\
	
	\While{$e<\mathrm{MaxEpoch}$}    
	{	
	$\mathcal{W} = \mathrm{BMM}(\mathcal{X},\mathcal{Y},\theta)$ \\
	
	\For{$b=1$ \KwTo $B$}
	{
	    $z_b = \mathrm{p_{model}}(x_b;\theta)$ \\
	    
	    $w_b$ = compute\_batch\_probs($x_b, \mathcal{W}, y_b$) \\
	    
	    $x^m_b, y^1_b, y^2_b, z^1_b, z^2_b, w^1_b, w^2_b, \lambda$ = mixup(\textbf{Augment}($x_b$), $y_b, z_b, w_b$) \\
	    
	    $z_m = \mathrm{p_{model}}(x^m_b;\theta)$ \\
	    $l_1 = (1 - w^1_b) \sum $NLLoss$(log(z_m), y^m_1) / |x_b| $\\
	    $l_2 = w^1_b \sum $NLLoss$(log(z_m), z^m_1) / |x_b| $\\
	    $l_3 = (1 - w^2_b) \sum $NLLoss$(log(z_m), y^m_2) / |x_b| $\\
	    $l_4 = w^2_b \sum $NLLoss$(log(z_m), z^m_2) / |x_b| $\\
	    
	    $\mathcal{L} = \lambda (l_1 + l_2) + (1 - \lambda) (l_3 + l_4) + \lambda_r \mathcal{L}_{reg}$\\
	    
    	$\theta=\mathrm{SGD}(\mathcal{L},\theta)$ \\
	}

}
	
{\bfseries Output $\theta$.}

\caption{\small Augmented M-DYR-H}
\label{alg:m-dyr-h}
	
\end{algorithm}

\subsection{Augmenting DivideMix}
A full version of the algorithm outlined in DivideMix is provided here (Algorithm \ref{alg:dividemix-full}). The technique is a combination of co-training, MixUp, loss modeling, and is trained in a semi-supervised learning manner. Notation and algorithm are as presented in the original paper \cite{li2020dividemix}. We insert our changes in bold. 

\begin{algorithm*}[t]
	
	\DontPrintSemicolon
	\small
	\textbf{Input:} $\theta^{(1)}$ and $\theta^{(2)}$, training dataset $(\mathcal{X},\mathcal{Y})$, clean probability threshold  $\tau$, number of augmentations $M$, augmentation policies Augment$_1$ and Augment$_2$, sharpening temperature $T$, unsupervised loss weight $\lambda_u$, $\mathrm{Beta}$ distribution parameter $\alpha$ for $\mathrm{MixMatch}$. \\

	$\theta^{(1)},\theta^{(2)}=\mathrm{WarmUp}(\mathcal{X},\mathcal{Y},\theta^{(1)},\theta^{(2)})$ \tcp*{standard training (with confidence penalty)}
	\While{$e<\mathrm{MaxEpoch}$}    
	{	
	$\mathcal{W}^{(2)} = \mathrm{GMM}(\mathcal{X},\mathcal{Y},\theta^{(1)})$ \tcp*{model per-sample loss with $\theta^{(1)}$ to obtain clean probability for $\theta^{(2)}$}
	$\mathcal{W}^{(1)} = \mathrm{GMM}(\mathcal{X},\mathcal{Y},\theta^{(2)})$ \tcp*{model per-sample loss with $\theta^{(2)}$ to obtain clean probability for $\theta^{(1)}$}

	\For( \tcp*[f]{train the two networks one by one}) {$k=1,2$}    
    {
	$\mathcal{X}^{(k)}_e = \{(x_i,y_i,w_i) | w_i\ge\tau, \forall (x_i,y_i,w_i) \in (\mathcal{X},\mathcal{Y},\mathcal{W}^{(k)})\}$\tcp*{labeled training set for $\theta^{(k)}$}
	$\mathcal{U}^{(k)}_e = \{x_i | w_i<\tau, \forall (x_i,w_i) \in (\mathcal{X},\mathcal{W}^{(k)})\}$  \tcp*{unlabeled training set for $\theta^{(k)}$}
    	
		\For {$\mathrm{iter}=1$ \KwTo $\mathrm{num\_iters}$}
		{
			From $\mathcal{X}^{(k)}_e$, draw a mini-batch $\{(x_b,y_b,w_b);b\in(1,...,B)\}$ \\
			From $\mathcal{U}^{(k)}_e$, draw a mini-batch $\{u_b;b\in(1,...,B)\}$ \\
			\For{$b=1$ \KwTo $B$}
			{
			    $x^{desc} = $ \textbf{Augment$_2$} $(x_b) $ \\
			    $u^{desc} = $\textbf{Augment$_2$} $(x_b) $ \\
			    
				\For{$m=1$ \KwTo $M$}
				{
					$\hat{x}_{b,m}=\mathrm{Augment_1}(x_b)$  \\
					$\hat{u}_{b,m}=\mathrm{Augment_1}(u_b)$ \\
				}
				${p}_{b}=\frac{1}{M}\sum_{m}\mathrm{p_{model}}(\hat{x}_{b,m};\theta^{(k)})$ \tcp*{average the predictions across augmentations of $x_b$}
				$\bar{y}_b=w_b y_b+(1-w_b){p}_{b}$ \\
				 	\tcp*{refine ground-truth label guided by the clean probability produced by the other network}
				$\hat{y}_b=\mathrm{Sharpen}(\bar{y}_b,T)$ 	\tcp*{apply temperature sharpening to the refined label}
				$\bar{q}_{b}=\frac{1}{2M}\sum_{m}\big(\mathrm{p_{model}}(\hat{u}_{b,m};\theta^{(1)})+\mathrm{p_{model}}(\hat{u}_{b,m};\theta^{(2)})\big)$\\
				 \tcp*{co-guessing: average the predictions from both networks across augmentations of $u_b$}
				
                $\hat{q_b} = Sharpen(\bar{q_b}, T)$ \\ 		\tcp*{apply temperature sharpening to the guessed label}		    
			}
            // train using a different augmentation \\
            $\hat{\mathcal{X}} = \{(x, y) | x \in x^{desc}, y \in \hat{y}\}$ \tcp*{\textbf{train with different augmentation}}
            $\hat{\mathcal{U}} = \{(u, q) | u \in u^{desc}, q \in \hat{q}\}$ \tcp*{\textbf{train with different augmentation}}
	        $\mathcal{L}_\mathcal{X},\mathcal{L}_\mathcal{U}=\mathrm{MixMatch}(\hat{\mathcal{X}},\hat{\mathcal{U}})$ \tcp*{apply MixMatch}
			$\mathcal{L}=\mathcal{L}_\mathcal{X}+\lambda_u \mathcal{L}_\mathcal{U}+\lambda_r \mathcal{L}_\mathrm{reg}$ \tcp*{total loss}
			$\theta^{(k)}=\mathrm{SGD}(\mathcal{L},\theta^{(k)})$  \tcp*{update model parameters}
		}
	}
	}
	\caption{\small Augmented DivideMix.}
	\label{alg:dividemix-full}
	
\end{algorithm*}

\vspace*{\fill}

\end{document}